\documentclass[runningheads]{llncs}
\usepackage[T1]{fontenc}
\usepackage{graphicx}
\usepackage{booktabs}
\usepackage[misc]{ifsym}
\newcommand{\corr}{(\Letter)}
\newcommand*\rot[1]{\rotatebox{90}{#1}}
\usepackage{amsmath, amssymb}
\usepackage{caption}
\usepackage[justification=centering]{subcaption}
\usepackage{pifont}
\newcommand{\cmark}{\ding{51}}
\newcommand{\xmark}{\ding{55}}
\usepackage{multirow}
\usepackage{xr-hyper}
\usepackage{mwe}

\begin{document}

\title{Joint-Centric Dual Contrastive Alignment with Structure-Preserving and Information-Balanced Regularization}


\titlerunning{Joint-Centric Dual Contrastive Alignment with Structure-Preserving} 


\author{Habibeh Naderi\inst{1} \corr \and
Behrouz Haji Soleimani\inst{1} \and
Stan Matwin\inst{1} 
}



\authorrunning{H. Naderi et al.}


\institute{Dalhousie University, Halifax NS, Canada \\ \email{habibeh.naderi@dal.ca, behrouz.hajisoleimani@dal.ca, stan@cs.dal.ca}
}

\maketitle              

\begin{abstract}
We propose HILBERT (HIerarchical Long-sequence Balanced Embedding with Reciprocal contrastive Training), a cross-attentive multimodal framework for learning document-level audio-text representations from long, segmented sequences in low-resource data settings. HILBERT leverages frozen pre-trained speech and language encoders to extract segment-level features, which are aggregated via cross-modal attention and self-attentive pooling to form modality-specific document representations and a joint cross-attentive embedding. To align modalities while preserving modality-specific structure under severe audio-text dimensional imbalance, we introduce a reciprocal dual contrastive objective that simultaneously aligns audio-to-joint and text-to-joint representations, rather than directly contrasting audio and text alone. Two auxiliary regularizers further stabilize long-sequence fusion: a Centered Kernel Alignment (CKA) loss that preserves structural consistency between each modality and the joint embedding, and a mutual information balancing loss that prevents dominance of a single modality by equalizing information flow from audio and text into the joint space. For downstream prediction, HILBERT employs a Mixture-of-Experts (MoE) classifier over concatenated audio, text, and joint representations to accommodate heterogeneous label regimes. Extensive evaluation across multiple audio-text backbone combinations demonstrates that HILBERT learns semantically meaningful long-sequence representations and achieves superior performance on highly imbalanced multi-class settings.

\end{abstract}

\section{Introduction}

Multimodal representation learning has emerged as a crucial research area \cite{VMoEs_shi2019variational}, leveraging the co-occurrence of observations from interdependent sources, such as paired audio and text, which act as a form of weak supervision. By integrating information across modalities, multimodal learning outperforms unimodal approaches in feature learning \cite{huang2024comparison}. However, effectively aligning representations from different modalities while preserving their distinctive characteristics remains a significant challenge \cite{crossCLR_zolfaghari2021crossclr}. Specifically, in audio-text multimodal learning, a key challenge arises from the high dimensionality of audio representations compared to text, potentially causing imbalanced contributions from each modality.

One of the key approaches for multimodal representation learning is contrastive learning. This method minimizes the distance between semantically related pairs while maximizing the distance between unrelated pairs in the embedded space. Through this objective, contrastive learning produces high-quality multimodal representations that exhibit robustness to distribution shifts and zero-shot transferability. CLIP exemplifies the success of contrastive learning in multimodal domains, playing a key role in advancing text-to-image generation techniques \cite{CLIP_radford2021learning}.

Simultaneously, sparsely activated Mixture-of-Experts (MoE) models \cite{stmoe_zloss_zoph2022st} have proven effective for expanding model capacity while maintaining manageable computational costs. By dynamically selecting a subset of parameters for each input, MoE models are particularly well-suited for multimodal learning since expert layers can learn an appropriate partitioning of modalities. These models enhance representation learning, improve multitask performance, and mitigate catastrophic forgetting in continual learning. The sparse MoE architecture provides several advantages: 1) increased model capacity without a proportional increase in computation, as only a subset of experts is typically active for any given input, 2) specialization of experts, where different experts can focus on different aspects of the data or different tasks, and 3) improved handling of heterogeneous data, which is common in multimodal scenarios.

In this work, we introduce HILBERT (HIerarchical Long-sequence Balanced Embedding with Reciprocal contrastive Training), a novel framework for multimodal audio-text representation learning that addresses the challenges of effective cross-modal alignment while preserving modality-specific information, particularly designed for long sequence representation. HILBERT balances modality contributions while preserving both shared and modality-specific features. Our approach leverages frozen pre-trained foundation models for feature extraction, employs a sophisticated dual contrastive learning strategy for cross-modal alignment, and utilizes an MoE architecture for downstream task learning. We design specialized loss functions including Centered Kernel Alignment (CKA) and Mutual Information (MI) losses to ensure balanced and informative joint representations. Our framework effectively handles the inherent imbalance between audio and text modalities, enabling more effective utilization of multimodal information for downstream tasks. HILBERT efficiently learns high-quality multimodal representations ensuring that the learned embeddings preserve semantic richness, effectively capturing both shared and modality-specific features.

The task of multimodal audio-text representation learning aims to map paired audio and text data into a joint representation space where semantically related pairs are placed close together, while unrelated pairs are pushed apart. A key challenge is balancing the contributions from different modalities, especially given the disparity in dimensionality and information density between audio and text data. Ensuring that the joint representation captures complementary information from both modalities, rather than being dominated by one, is crucial for effective multimodal learning. HILBERT incorporates a dual multimodal contrastive loss, designed to preserve structural alignment between modality-specific and joint representations. Unlike conventional contrastive methods that primarily focus on inter-modality alignment, our approach enforces both inter- and intra-modality consistency, preventing semantically similar representations from being pushed apart in the embedding space. Inspired by prior contrastive learning frameworks such as SimCLR \cite{simclr_chen2020simple} and CrossCLR \cite{crossCLR_zolfaghari2021crossclr}, HILBERT extends contrastive learning principles to multimodal data, improving the quality of learned joint embeddings \cite{gmc_poklukar2022geometric}.

HILBERT targets a learning regime that is fundamentally distinct from CLAP-style \cite{elizalde2023clap} and related cross-modal pretraining frameworks. CLAP variants are designed for large-scale audio-caption corpora and optimized for short audio clips through global embedding alignment, making them ill-suited for long, document-length audio-text inputs. In contrast, HILBERT is explicitly developed for long-sequence, document-level representation learning, leveraging cross-modal self-attention to model segment-level interactions, auxiliary CKA and mutual information losses to balance modality contributions, and an MoE architecture for multi-task prediction. Unlike CLAP's contrastive objective, which operates on pooled audio and text embeddings and does not capture long-range temporal structure, HILBERT directly models cross-modal dependencies across extended sequences. We therefore view CLAP and similar large-scale pretraining methods as complementary, while HILBERT provides a lightweight, backbone-agnostic framework for segment-aware audio-text integration under small-data and constrained training conditions.

By integrating frozen pre-trained models, contrastive objectives, and MoE-based adaptation, HILBERT advances the state-of-the-art in multimodal representation learning for long sequences. Our experiments demonstrate that HILBERT achieves competitive performance in capturing semantically meaningful representations, making it well-suited for a wide range of downstream tasks.

The main contributions of this work are as follows:
\begin{itemize}
    \item We introduce a dual contrastive learning strategy that simultaneously enforces cross-modal and modality-specific consistency. Unlike conventional contrastive methods that primarily focus on inter-modality alignment, our approach explicitly preserves intra-modality structure, preventing semantically similar instances from being separated in the embedding space.
    
    \item We incorporate specialized loss functions including CKA and MI losses to enhance cross-modal alignment while preserving information-theoretic relevance. CKA loss ensures inter-modal and intra-modal alignment while MI loss ensures equitable contribution from both modalities to the joint representation. These specialized loss functions improve the quality of learned representations by ensuring that the joint embedding space remains semantically meaningful and well-structured.
    
    \item We develop a multimodal joint encoder architecture that effectively combines frozen pre-trained foundation models with cross-modal self-attention mechanisms to capture complex dependencies between audio and text modalities. This multi-head self-attention mechanism generates high quality document-level representations from segment-level embeddings, capturing complex dependencies among segments.
    
    \item We integrate an MoE approach for downstream task learning that adaptively leverages information from different experts based on input features, enabling efficient specialization across heterogeneous data distributions.
\end{itemize}

\section{Proposed Method}
In this section, we present our novel framework for multimodal audio-text representation learning, which we call HILBERT (HIerarchical Long-sequence Balanced Embedding with Reciprocal contrastive Training). Our approach leverages pre-trained models for effective representation learning and addresses the challenge of alignment in multimodal learning. The proposed framework extends the principles of multimodal contrastive representation learning and builds on prior state-of-the-art \cite{gmc_poklukar2022geometric,pmi_shi2020relating} while introducing several innovations to enhance representation quality and cross-modal alignment.

The task of multimodal audio-text representation learning aims to learn effective joint representations from paired audio and text data. Let $\mathcal{D} = \{(x_i^a, x_i^t, y_i)\}_{i=1}^N$ denote a dataset containing $N$ audio-text pairs, where $x_i^a$ represents the audio input, $x_i^t$ denotes the corresponding text input, and $y_i$ is the associated label (e.g., emotion, speaker identity, or semantic category). Our objective is to learn a joint representation space where semantically related audio and text pairs are mapped close to each other, while unrelated pairs are pushed apart.

More formally, we aim to train a set of encoders that can map audio inputs $x^a$ and text inputs $x^t$ into a shared latent space, ensuring that the joint representation effectively captures complementary information from both modalities. This embedding space preserves high-quality modality-specific features while also modeling cross-modal interactions, making it well-suited for downstream tasks.
\begin{figure}[!t]
	\centering
    \includegraphics[width=0.6\columnwidth]{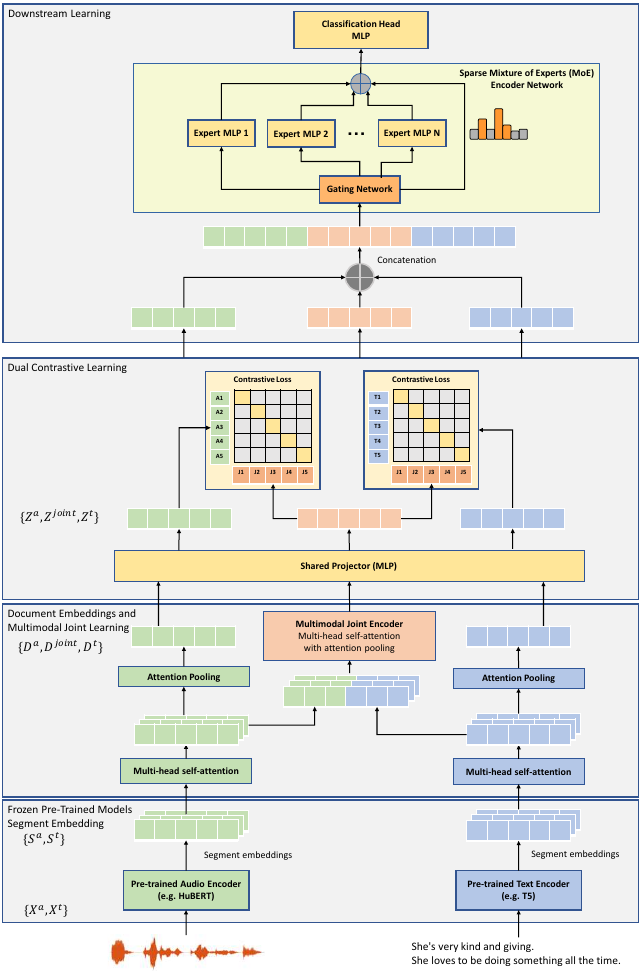}
    \caption{\label{fig_model_architecture}Our proposed HILBERT model architecture.}
\end{figure}

\subsubsection{\textbf{HILBERT Model Architecture:}}
Our HILBERT framework comprises four main components as illustrated in Figure \ref{fig_model_architecture}. This architecture is specifically designed to process long sequences, such as large documents and extended audio files. It includes a segmentation step, which can be performed manually or automatically, to break the input into smaller segments before passing them to the model.

\begin{enumerate}
\item	\textbf{Frozen Pre-trained Models and Segment Embedding:} We utilize pre-trained models for audio and text to extract rich feature representations from raw inputs. These pre-trained models have been trained on large-scale datasets and can provide high-quality embeddings that capture semantic and structural information.
\item	\textbf{Document Embeddings and Multimodal Joint Learning:} The segment-level embeddings from the pre-trained models are processed through a multi-head self-attention mechanism to generate document-level representations. A multimodal joint encoder then combines information from both modalities.
\item	\textbf{Dual Contrastive Learning:} This step consists of a shared encoder that maps inputs into a joint representation. We employ a contrastive learning approach that involves contrasting positive pairs (semantically related audio-text pairs) against negative pairs (unrelated pairs). Our designed contrastive loss incorporates multimodal knowledge into the each modality embeddings and aligns representations from different modalities through a Centered Kernel Alignment (CKA) loss. To further enhance relevance, we introduce a Mutual Information (MI) loss, ensuring the joint embedding retains meaningful information from each modality. 
\item	\textbf{Downstream Learning with Mixture of Experts (MoE):} For downstream tasks, we utilize an MoE architecture to adaptively leverage information from different experts based on the input.
\end{enumerate}

Our pipeline is hierarchical: long audio and text are segmented, encoded with frozen models, and aggregated via document-level multi-head self-attention with modality-specific pooling and cross-modal fusion. This reduces complexity from frame-level $O(T^2)$ to segment-level $O(L^2)$, with $L$ in the tens. Only the segment-to-document and fusion layers are trained, keeping compute bounded while preserving long-range dependencies. In the following subsections, we elaborate on each component and discuss how they contribute to effective multimodal representation learning.

\subsection{Frozen Pre-trained Models and Feature Extraction}
For audio encoding, we utilize pre-trained models (e.g., Whisper, HuBERT) that has been trained on large-scale audio data. The pre-trained audio encoder $f_E^a(\cdot)$ extracts rich acoustic and semantic features from the raw audio signal. Given an audio input $\mathbf{X}^a$, the pre-trained encoder processes it to generate segment-level embeddings, $\mathbf{S}^a = f_E^a(\mathbf{X}^a) \in \mathbb{R}^{L \times d_a}$, where $L$ is the number of segments, and $d_a$ is the dimension of the audio embedding space.

Similarly, for text encoding, we employ pre-trained language models (e.g., T5, RoBERTa) that has been trained on large-scale text corpora. The pre-trained text encoder $f_E^t(\cdot)$ processes the text input $\mathbf{X}^t$ to generate segment-level embeddings, $\mathbf{S}^t = f_E^t(\mathbf{X}^t) \in \mathbb{R}^{L \times d_t}$, where $L$ again is the number of segments, and $d_t$ is the dimension of the text embedding space. These segment-level embeddings capture semantic information from the text.

It is important to note that the parameters of these pre-trained models are kept frozen during training to avoid the computational overhead associated with fine-tuning large models and to prevent overfitting on limited data. This approach aligns with the philosophy of leveraging the rich knowledge encoded in pre-trained models while keeping the training process efficient.

\subsection{Document Embeddings and Multimodal Joint Learning}

\subsubsection{\textbf{Modality-Specific Document Embedding:}}
We first project segment-level embeddings of audio and text (i.e. $\mathbf{S}^a_{d_a}, \mathbf{S}^t_{d_t}$ ) to the same dimension to obtain $\mathbf{S}^a_d, \mathbf{S}^t_d$. Then to obtain document-level representations for audio and text (i.e. $\mathbf{D}^{a}$, $\mathbf{D}^{t}$) from their corresponding segment-level embeddings, we employ multi-head self-attention mechanisms with attention pooling. The multi-head self-attention mechanism allows the model to attend to different parts of the input sequence and capture complex dependencies and relationships among segments. 

\subsubsection{\textbf{Multimodal Joint Encoder:}}
The multimodal joint learning module contains a joint encoder $f_E^{joint}(\cdot)$ designed to efficiently fuse the audio-text information. It consists of a cross-modal fusion layer and follows attention-based embedding layers. To achieve this, we take the audio-to-text embedding $\mathbf{S}^{a \to t}$ to show the cross-modal attention:
\begin{equation}
\small{
    \mathbf{S}^{a \to t} = CM_{a \to t} (\mathbf{S}^a_d, \mathbf{S}^t_d) = \text{softmax} \left( \frac{W_{Q_a} \mathbf{S}^a_d W_{K_t}^T {\mathbf{S}^t_d}^T}{\sqrt{d}} \right) W_{V_t}
}
\end{equation}
Similarly, we can get text-to-audio embedding $\mathbf{S}^{t \to a}$. The obtained $\mathbf{S}^{a \to t}$ and $\mathbf{S}^{t \to a}$ will be concatenated together and projected to the latent space as the joint document embedding via a self-attention layer followed by attention pooling:
\begin{equation}
    \mathbf{D}^{joint} = f_{AttPool}(\text{SelfAttention} (\mathbf{S}^{a \to t} \oplus \mathbf{S}^{t \to a}))
\end{equation}

The multimodal joint encoder combines information from both modalities and captures cross-modal interactions. The design of the joint encoder is crucial for effective multimodal representation learning. Unlike simple concatenation or averaging, our joint encoder employs a more sophisticated fusion mechanism that can adaptively weigh the contributions from different modalities based on their informativeness and reliability.

\subsection{Contrastive Learning and Cross-Modal Alignment}

To enhance the alignment between different modalities and promote effective joint representation learning, we employ a contrastive learning approach. Contrastive learning aims to bring the representations of semantically related samples closer while pushing apart the representations of unrelated samples in the embedding space. This module consists of a shared projector to map all inputs to the same latent space. Given the presence of a joint encoder in our architecture, we employ a dual contrastive learning strategy, which enforces alignment between (i) audio and joint representations and (ii) text and joint representations. Our specialized dual contrastive loss function is composed of three key components: 1) contrastive loss, 2) Centered Kernel Alignment (CKA) loss, and 3) Mutual Information (MI) loss.

\subsubsection{\textbf{Shared Projector:}}
The document-level representations $\mathbf{D}^a$ and $\mathbf{D}^t$ from both modalities as well as the joint document-level embedding $\mathbf{D}^{joint}$ are then fed into a shared projector implemented as a multilayer perceptron (MLP) to map them into a common latent space, $\mathbf{Z}^a = f_{shared}(\mathbf{D}^a) \in \mathbb{R}^{d_{latent}}$, $\mathbf{Z}^t = f_{shared}(\mathbf{D}^t) \in \mathbb{R}^{d_{latent}}$, $\mathbf{Z}^{joint} = f_{shared}(\mathbf{D}^{joint}) \in \mathbb{R}^{d_{latent}}$ where $d_{latent}$ is the dimension of the common latent space and $f_{shared}(\cdot)$ is the shared projection function.

\subsubsection{\textbf{Multimodal Contrastive Loss:}}
We utilize a multimodal contrastive loss to align the joint representation with modality-specific representations. Given a batch of embeddings $B = \{Z_i^a, Z_i^t, Z_i^{joint}\}_{i=1}^{|B|}$, the positive pairs are defined as the joint embedding with its corresponding modality-specific embeddings, i.e., $(Z_i^a, Z_i^{joint})$ and $(Z_i^t, Z_i^{joint})$. All other pairing combinations are treated as negative pairs.

Let $\text{c}(x_i, x_j)$ represent the cosine similarity between two embeddings $x_i$ and $x_j$, and $\tau$ be the temperature hyperparameter. The scaled similarity is defined as $\text{sim}(x_i, x_j) = \exp\left(\frac{\text{c}(x_i, x_j)}{\tau}\right)$ and the negative pairs for contrastive learning are defined as:
\begin{equation}
\Omega_i^m = \sum_{j \neq i} \left(\text{sim}(Z_i^m, Z_j^m) + \text{sim}(Z_i^m, Z_j^{joint}) + \text{sim}(Z_i^{joint}, Z_j^{joint})\right)
\end{equation}
where $m \in \{a, t\}$ indicates the modality type. The contrastive loss for all data embeddings is then formulated as:
\begin{equation}
L_{con}(B) = -\frac{1}{|B|} \sum_{i=1}^{|B|} \log\left(\frac{\text{sim}(Z_i^a, Z_i^{joint})}{\Omega_i^a} + \frac{\text{sim}(Z_i^t, Z_i^{joint})}{\Omega_i^t}\right)
\end{equation}
This loss encourages the joint representation to be semantically aligned with both the audio and text representations of the same sample while being distinct from the representations of other samples. Unlike pairwise audio-text, modality-modality, joint-joint schemes, we deliberately avoid audio-text contrast in the objective and instead anchor both modalities to the fused joint space, then enforce covariance alignment and information balance. This reduces modality collapse without over-penalizing intra-modality neighbors, and is tailored to high-dimensional audio versus compact text and to our long-sequence hierarchical fusion.

\subsubsection{\textbf{Centered Kernel Alignment (CKA) Loss:}}
We propose to use Centered Kernel Alignment (CKA) as a measure of similarity between representation spaces. CKA is a powerful tool for measuring the similarity between neural network representations and has been shown to be invariant to orthogonal transformations and isotropic scaling \cite{cka_kornblith2019similarity}. Given two sets of representations $X$ and $Y$, the CKA between them is defined as:
\begin{equation}
\small{
\text{CKA}(K, L) = \frac{\text{HSIC}(K, L)}{\sqrt{\text{HSIC}(K, K) \cdot \text{HSIC}(L, L)}}
}
\end{equation}
where $K = XX^T$ and $L = YY^T$ are the kernel matrices, and HSIC (Hilbert-Schmidt Independence Criterion) \cite{hsic_gretton2005measuring} is defined as $\text{HSIC}(K, L) = \frac{1}{(n-1)^2} \text{trace}(KHLH)$ where $n$ is the number of samples, $H = I_n - \frac{1}{n} \mathbf{1}_n \mathbf{1}_n^T$ is the centering matrix, $I_n$ is the identity matrix, and $\mathbf{1}_n$ is a vector of ones of length $n$. Given the centered data matrices $\tilde{X} = HX$ and $\tilde{Y} = HY$, we can express the HSIC in terms of covariance:
\begin{equation}
\text{HSIC}(K, L) = \frac{1}{(n-1)^2} \text{trace}(\tilde{X}\tilde{X}^T \tilde{Y}\tilde{Y}^T) = \|\text{Cov}(\tilde{X}^T, \tilde{Y}^T)\|_F^2
\end{equation}
where $\text{Cov}(X,Y)$ is the covariance matrix and $\|\cdot\|_F^2$ is the Frobenius norm of the matrix. With these definitions, the CKA can be formulated as:
\begin{equation}
\small{
\text{CKA}(X, Y) = \frac{\|\text{Cov}(X, Y)\|_F^2}{\|\text{Cov}(X, X)\|_F \cdot \|\text{Cov}(Y, Y)\|_F}
}
\end{equation}
CKA ranges between 0 and 1, therefore for our multimodal scenario, the CKA loss components become:
\begin{equation}
L_{CKA}^m = 1 - \frac{\|\text{Cov}({Z^{joint},Z^m})\|_F^2}{\|\text{Cov}(Z^{joint}, Z^{joint})\|_F \cdot \|\text{Cov}(Z^m, Z^m)\|_F}
\end{equation} 
where $m\in\{a, t\}$ is the modality. The total CKA loss remains:
\begin{equation}
L_{CKA} = L_{CKA}^a + L_{CKA}^t
\end{equation}


This covariance formulation highlights that CKA measures the normalized alignment between the covariance structures of the representations, providing deeper insight into how information is preserved across modalities in the joint representation.

\subsubsection{\textbf{Mutual Information (MI) Loss:}}
To further enhance the informativeness of the joint representation, we introduce an MI loss that maximizes the mutual information between the joint representation and each modality-specific representation. Mutual information measures the amount of information obtained about one random variable through observing another random variable.

Since direct computation of mutual information is challenging, we adopt the InfoNCE estimator, which provides a lower bound on the mutual information:
\begin{equation}
I(Z^{joint}; Z^m) \geq \mathbb{E}\left[\frac{1}{|B|} \sum_{i=1}^{|B|} \log \frac{f(Z_i^{joint}, Z_i^m)}{\frac{1}{|B|} \sum_{j=1}^{|B|} f(Z_i^{joint}, Z_j^m)}\right] = L_{MI}^m
\end{equation}
where $f(x, y)$ is a scoring function that measures the compatibility between $x$ and $y$. We use the dot product as the scoring function, $f(x, y) = \exp(x^T y / \gamma)$ where $\gamma$ is a temperature parameter. The total MI term balances contributions of modalities in the joint embedding and prevents over-emphasis on a single modality:
\begin{equation}
L_{MI} = -(L_{MI}^a + L_{MI}^t) + (L_{MI}^a - L_{MI}^t)^2
\end{equation}
By minimizing the difference between the mutual information of the joint representation and each modality-specific representation, we ensure that the joint representation captures the essential information from both modalities.

HILBERT uses CKA and MI as mechanisms for modality balancing, not for forcing audio and text to share identical covariance structure. The joint encoder, via cross-attention, naturally integrates both modalities, CKA encourages structural consistency between each modality and the joint embedding, while MI prevents dominance of a single modality. In HILBERT, the modality gap is addressed in three ways: 1) A shared projector maps both modality-specific and joint document embeddings into a common latent space, encouraging a comparable representation scale. 2) CKA between each modality and the joint embedding (not audio$\leftrightarrow$text directly), ensuring structural information from both inputs is retained without forcing identical covariance. This reduces representational mismatch while preserving modality-specific structure. 3) The MI term explicitly balances the mutual information between the joint embedding and each modality, penalizing dominance of one modality and promoting equitable information flow and stabilizing long-sequence fusion. This is related in spirit to Cacophony's \cite{zhu2024cacophony} modality-balancing goals but integrated into a long-sequence cross-attentive encoder and MoE classifier.

\subsection{Downstream Learning with Mixture of Experts (MoE)}
For downstream tasks such as classification, we propose an MoE architecture that can adaptively leverage information from different experts based on the input. The MoE architecture consists of multiple expert networks and a gating network that determines the contribution of each expert to the final output.

Each expert network is implemented as a multilayer perceptron (MLP) that takes the joint representation as input and produces a task-specific output $E_i(Z) = f_{MLP}^i(Z)$ where $Z=[Z^a ; Z^{joint} ; Z^t]$ represents the concatenation of audio, text, and joint features. The gating network determines the contribution of each expert to the final output. It takes the concatenated features from all modalities as input and outputs a probability distribution over the experts $g = f_{gate}(Z) \in \mathbb{R}^{N_E}$ where $N_E$ is the number of experts, and $f_{gate}(\cdot)$ is implemented as a neural network with softmax activation to ensure that the gating values sum to 1: 
\begin{equation}
g_i = \frac{\exp(f_{gate}^i(Z))}{\sum_{j=1}^{N_E} \exp(f_{gate}^j(Z))}
\end{equation}

The gating network enables the model to dynamically select the most relevant experts based on the input features. This is particularly useful in multimodal scenarios where different experts may specialize in different aspects of the data. The outputs from all experts are combined using the gating values to produce the final representation, $Z_{MoE} = \sum_{i=1}^{N_E} g_i \cdot E_i(Z)$. This weighted combination allows the model to adaptively leverage the strengths of different experts based on the input.

For classification tasks, the MoE output is passed through a final classification head implemented as an MLP, $\hat{y} = f_{sup}(Z_{MoE}) \in \mathbb{R}^{C}$ where $C$ is the number of classes, and $f_{sup}(\cdot)$ is a multilayer perceptron with softmax activation. For the supervised classification loss, we use the cross-entropy loss:
\begin{equation}
L_{sup} = -\frac{1}{|B|} \sum_{i=1}^{|B|} \sum_{c=1}^{C} y_{i,c} \log(\hat{y}_{i,c})
\end{equation}
where $y_{i,c}$ is a binary indicator if class label $c$ is the correct classification for sample $i$, and $\hat{y}_{i,c}$ is the predicted probability that sample $i$ belongs to class $c$.

\subsubsection{\textbf{Total Loss Function:}}
The overall loss function for our multimodal representation learning framework combines the supervised classification loss, contrastive loss, CKA loss, and MI loss:
\begin{equation}
L_{total} = L_{sup} + \lambda_{con} L_{con} + \lambda_{CKA} L_{CKA} + \lambda_{MI} L_{MI}
\end{equation}
where $L_{sup}$ is the supervised classification loss (e.g., cross-entropy loss for classification tasks), and $\lambda_{con}$, $\lambda_{CKA}$, and $\lambda_{MI}$ are hyperparameters that control the contribution of each loss term.

\section{Dataset}
\label{sec_data}

The data used in this work consists of audio speech samples from 369 subjects participating in the Families Overcoming Risks and Building Opportunities for Well Being (FORBOW) research project \cite{forbow_uher2014familial}. Participants are parents, 266 mothers and 103 fathers, in the age range of 28-51 years. In these clinical interviews, parents were asked to talk about their children for five minutes without interruption. Out of these subjects, 149 were diagnosed with Major Depressive Disorder (MDD), 66 were diagnosed with Bipolarity Disorder (BD), 19 were diagnosed with Schizophrenia, and 129 were the control group with no major mood disorders. In addition to the parents' interview files, FORBOW research project started collecting interviews with the children themselves. The audio interviews of children consists of 3 parts: 1) a three minute interview where children talk about themselves, 2) a two minute interview talking about a positive experience they had, and 3) a two minute interview where they talk about a negative experience they had. All three interviews are uninterrupted with a total of 7 minutes of speech from each child. 

We transcribed and broke down each sample into multiple segments based on changes in emotion, sentiment, objectivity/subjectivity, etc. Average word count in a segment is 17 and average audio length for a segment is 6.47 seconds.Table \ref{tbl_dataset_info} summarizes the dataset statistics, including the set of prediction tasks at different levels of granularity (segment-level, document-level, and psychological and cognitive tasks), as well as the label distributions and the degree of class imbalance.

\begin{table}[!t]
\centering
\caption{Dataset information for the parent and offspring data.}
\label{tbl_dataset_info}
\setlength\tabcolsep{3pt}
\scalebox{0.65}{
\begin{tabular}{lllcccccccc}
\toprule
& \textbf{Level} & \textbf{Task} & \rot{\textbf{Num Samples}} & \rot{\textbf{Num Classes}} & \rot{\textbf{Imbalanced (\%)}} & \rot{\textbf{Label 0}} & \rot{\textbf{Label 1}} & \rot{\textbf{Label 2}} & \rot{\textbf{Label 3}} & \rot{\textbf{Label 4}} \\
\midrule
Parent & Document & affect & 484 & 3 & 93.18 & 23 & 124 & 337 & - & - \\
Parent & Document & warmth & 485 & 3 & 39.69 & 117 & 194 & 174 & - & - \\
Parent & Document & overprotection & 487 & 3 & 85.77 & 267 & 182 & 38 & - & - \\
Parent & Document & cohesion & 492 & 5 & 92.63 & 14 & 14 & 132 & 190 & 142 \\
Parent & Document & criticism & 492 & 4 & 89.22 & 232 & 166 & 69 & 25 & - \\
Parent & Document & worry & 381 & 4 & 93.29 & 164 & 147 & 59 & 11 & - \\
\midrule
Parent & Cognitive & spectrum & 363 & 4 & 87.25 & 129 & 149 & 66 & 19 & - \\
Parent & Cognitive & introvert & 369 & 2 & 49.39 & 245 & 124 & - & - & - \\
Parent & Cognitive & extrovert & 369 & 2 & 63.84 & 271 & 98 & - & - & - \\
Parent & Cognitive & ADHD & 369 & 2 & 67.74 & 279 & 90 & - & - & - \\
Parent & Cognitive & anxiety & 369 & 2 & 51.81 & 249 & 120 & - & - & - \\
Parent & Cognitive & depression & 369 & 2 & 89.19 & 333 & 36 & - & - & - \\
\midrule
Offspring & Document & affect & 621 & 4 & 97.73 & 16 & 11 & 109 & 485 & - \\
Offspring & Document & coherence & 616 & 5 & 95.83 & 11 & 33 & 109 & 264 & 199 \\
Offspring & Document & richness & 614 & 3 & 96.70 & 82 & 515 & 17 & - & - \\
\midrule
Offspring & Cognitive & spectrum & 85 & 4 & 86.11 & 36 & 27 & 17 & 5 & - \\
Offspring & Cognitive & introvert & 85 & 2 & 39.62 & 53 & 32 & - & - & - \\
Offspring & Cognitive & extrovert & 85 & 2 & 53.45 & 58 & 27 & - & - & - \\
Offspring & Cognitive & ADHD & 85 & 2 & 53.45 & 58 & 27 & - & - & - \\
Offspring & Cognitive & anxiety & 85 & 2 & 39.62 & 53 & 32 & - & - & - \\
Offspring & Cognitive & depression & 85 & 2 & 83.56 & 73 & 12 & - & - & - \\
\bottomrule
\end{tabular}}
\end{table}





\section{Experimental Results}

In this section, we provide the experimental results of our proposed contrastive learning framework. We chose six different language models, nliRoBERTa, nliDistilRoBERTa, nliMPNet, paraTinyBERT, sentenceT5XL, and allMiniLM12. We also chose five pre-trained audio models, whisperMedium, wav2vec2Large-FineTune, hubertLargeFineTune, conformerLargeFineTune, and spectrogram. This results in a total of 30 combinations, trained separately on each individual task for both parent and offspring data.

For the dimensionality of the shared projector in contrastive embedding we tested 64, 128, and 256 and the quality of predictions was not sensitive to the dimensionality of the contrastive latent space. Therefore, in the following we only report the results achieved with dimensionality of 128. The expert network consists of 8 experts each of which is a 2 layer MLP with 32 and 32 units. The classification head consists of 3 layer dense network, a 32-unit layer followed by a 16-unit layer followed by a softmax layer with the size corresponding to the classification task. All results in this section are based on a 25-fold cross-validation.

\begin{table*}[!t]
\centering
\caption{AUC (\%) on a 25 fold cross-validation for document-level and psychological spectrum predictions using parent data.}
\label{tbl_auc_parent}
\setlength\tabcolsep{2.75pt}
\makebox[\textwidth][c]{
\scalebox{0.7}{
\begin{tabular}{l|cccccc|ccccccc}
\toprule
& \multicolumn{6}{|c|}{\textbf{Document-level tasks}} & \multicolumn{7}{|c}{\textbf{Cognitive and Psychological tasks}} \\
\midrule
 & \rot{\textbf{affect}} & \rot{\textbf{warmth}} & \rot{\textbf{overprotect}} & \rot{\textbf{cohesion}} & \rot{\textbf{criticism}} & \rot{\textbf{worry}} & \rot{\textbf{Spectrum}} & \rot{\textbf{Introvert}} & \rot{\textbf{Extrovert}} & \rot{\textbf{ADHD}} & \rot{\textbf{Anxiety}} & \rot{\textbf{Depression}} & \rot{\textbf{Mood}} \\
 &  &  &  &  &  &  &  &  &  &  &  &  &  \\
\midrule
\textbf{CLAP-LAION \cite{wu2023large_clapstyle}} & 71.84 & 56.11 & 51.73 & 60.65 & 56.96 & 54.86 & 53.09 & 55.11 & 56.18 & 54.07 & 57.03 & 59.86 & 57.51 \\
\midrule
\textbf{TinyBERT+hubLgFT (Baseline: Transfer)} & 72.88 & 59.67 & 57.21 & 63.93 & 57.02 & 59.33 & 58.53 & 61.50 & 56.77 & 57.11 & 57.61 & 59.20 & 63.60 \\
\textbf{TinyBERT+hubLgFT (Transfer + MoE)} & 73.12 & 59.21 & 57.49 & 63.70 & 58.16 & 60.88 & 58.44 & 61.76 & 57.59 & 55.85 & 58.82 & 60.24 & 64.13 \\
\textbf{TinyBERT+hubLgFT (Contrastive + MoE)} & 74.68 & 61.98 & 58.89 & 64.91 & 59.41 & 61.72 & 59.47 & 63.09 & 59.35 & 58.88 & 60.18 & 61.77 & 65.61 \\
\textbf{TinyBERT+hubLgFT (HILBERT: DualC + MoE)} & 76.30 & 62.70 & 57.51 & 67.67 & \textbf{61.98} & \textbf{63.02} & 61.20 & \textbf{65.38} & 59.05 & 59.87 & 62.82 & 61.85 & 67.63 \\
\midrule
\textbf{TinyBERT+confLgFT (Baseline: Transfer)} & 71.17 & 60.73 & 50.65 & 60.82 & 55.31 & 53.75 & 62.78 & 55.23 & 59.67 & 58.99 & 58.50 & 58.17 & 64.99 \\
\textbf{TinyBERT+confLgFT (Transfer + MoE)} & 72.18 & 61.75 & 52.44 & 62.10 & 56.11 & 55.00 & 64.43 & 55.31 & 59.94 & 59.52 & 59.77 & 58.76 & 65.15 \\
\textbf{TinyBERT+confLgFT (Contrastive + MoE)} & 74.90 & 64.31 & 53.30 & 63.83 & 57.26 & 56.68 & 66.11 & 56.26 & 62.05 & 61.14 & 61.84 & 60.81 & 67.10 \\
\textbf{TinyBERT+confLgFT (HILBERT: DualC + MoE)} & 74.85 & 66.54 & 54.91 & 63.46 & 60.37 & 57.41 & \textbf{66.75} & 58.16 & 65.14 & 61.52 & 63.57 & 63.00 & 69.62 \\
\midrule
\textbf{nRoBERTa+hubLgFT (Baseline: Transfer)} & 70.31 & 61.50 & 54.31 & 59.33 & 55.93 & 59.88 & 53.80 & 56.02 & 58.97 & 54.92 & 60.44 & 59.88 & 66.05 \\
\textbf{nRoBERTa+hubLgFT (Transfer + MoE)} & 71.91 & 61.32 & 56.29 & 59.34 & 56.76 & 58.55 & 55.47 & 57.53 & 59.64 & 55.21 & 60.98 & 62.66 & 67.68 \\
\textbf{nRoBERTa+hubLgFT (Contrastive + MoE)} & 73.43 & 63.40 & 56.99 & 61.76 & 58.61 & 60.60 & 56.01 & 59.07 & 60.51 & 58.10 & 62.47 & 63.65 & 69.31 \\
\textbf{nRoBERTa+hubLgFT (HILBERT: DualC + MoE)} & 75.30 & 64.59 & 57.61 & 63.52 & 58.92 & 58.60 & 58.21 & 61.85 & 59.21 & 58.49 & \textbf{64.89} & \textbf{67.87} & \textbf{70.99} \\
\midrule
\textbf{TinyBERT+w2v2LgFT (Baseline: Transfer)} & 71.80 & 60.26 & 53.54 & 63.11 & 55.61 & 53.30 & 58.40 & 58.12 & 61.34 & 56.44 & 56.46 & 59.85 & 63.88 \\
\textbf{TinyBERT+w2v2LgFT (Transfer + MoE)} & 72.99 & 59.76 & 55.03 & 65.54 & 56.06 & 54.62 & 60.64 & 60.25 & 60.57 & 56.33 & 58.14 & 59.01 & 63.96 \\
\textbf{TinyBERT+w2v2LgFT (Contrastive + MoE)} & 73.84 & 61.87 & 55.60 & 65.40 & 58.05 & 56.68 & 61.18 & 60.69 & 62.69 & 58.37 & 60.07 & 61.79 & 66.39 \\
\textbf{TinyBERT+w2v2LgFT (HILBERT: DualC + MoE)} & 74.58 & 63.29 & 58.97 & 66.24 & 58.70 & 57.73 & 61.64 & 62.49 & 66.37 & 60.42 & 61.40 & 63.08 & 66.09 \\
\midrule
\textbf{nMPNet+hubLgFT (Baseline: Transfer)} & 74.47 & 63.98 & 56.39 & 63.20 & 54.30 & 55.12 & 55.75 & 57.19 & 59.94 & 57.13 & 56.29 & 58.01 & 60.38 \\
\textbf{nMPNet+hubLgFT (Transfer + MoE)} & 74.71 & 63.38 & 56.04 & 63.14 & 56.71 & 55.07 & 56.77 & 56.74 & 60.78 & 58.32 & 56.78 & 59.78 & 60.36 \\
\textbf{nMPNet+hubLgFT (Contrastive + MoE)} & 77.21 & 65.69 & 57.77 & 64.65 & 57.17 & 57.57 & 57.68 & 58.61 & 62.52 & 59.48 & 59.38 & 61.17 & 61.93 \\
\textbf{nMPNet+hubLgFT (HILBERT: DualC + MoE)} & 79.96 & \textbf{67.24} & \textbf{59.19} & \textbf{70.29} & 59.19 & 58.28 & 58.80 & 61.70 & 64.31 & 59.33 & 62.18 & 61.98 & 65.92 \\
\midrule
\textbf{nRoBERTa+w2v2LgFT (Baseline: Transfer)} & 73.98 & 59.61 & 48.47 & 62.92 & 58.31 & 54.19 & 51.81 & 57.82 & 64.94 & 57.08 & 55.71 & 59.89 & 63.68 \\
\textbf{nRoBERTa+w2v2LgFT (Transfer + MoE)} & 73.97 & 62.21 & 49.31 & 61.92 & 58.87 & 56.54 & 52.47 & 58.41 & 64.94 & 57.59 & 57.16 & 61.59 & 63.96 \\
\textbf{nRoBERTa+w2v2LgFT (Contrastive + MoE)} & 76.11 & 63.40 & 50.85 & 64.83 & 59.33 & 56.80 & 54.66 & 60.42 & 66.41 & 59.02 & 58.33 & 61.68 & 65.80 \\
\textbf{nRoBERTa+w2v2LgFT (HILBERT: DualC + MoE)} & \textbf{80.34} & 63.72 & 52.53 & 68.25 & 60.88 & 59.02 & 54.98 & 62.61 & \textbf{67.83} & \textbf{61.90} & 61.46 & 66.57 & 68.17 \\
\bottomrule
\end{tabular}}
}
\end{table*}

Table \ref{tbl_auc_parent} reports AUC (\%) from 25-fold cross-validation over 6 backbone combinations and 4 architecture configurations on 6 document-level and 7 psychological spectrum tasks using parent data, with methods sorted by average performance across all tasks. Overall, the results demonstrate a clear and consistent advantage of the full HILBERT (Dual Contrastive + MoE) framework across nearly all backbone combinations. Adding MoE to transfer learning yields modest but systematic improvements, while introducing contrastive learning leads to further gains, particularly on more challenging psychological traits. The complete HILBERT configuration achieves the strongest performance overall, with pronounced improvements on document-level affective tasks and gains of up to 5-10 AUC points over CLAP-based pre-training, highlighting the benefit of task-aware multimodal contrastive learning over generic audio-text alignment.

Across backbones, nMPNet and nRoBERTa combined with hubERT or w2v2 achieve the highest absolute AUCs (e.g., $\sim80$ for affect and $\sim70$ for cohesion/mood), indicating that stronger text encoders better leverage dual-contrastive supervision. Performance gains are consistent across both task families, suggesting improved generalization rather than task-specific overfitting. Notably, TinyBERT+ConformerLarge (HILBERT) attains the best result on the spectrum task (66.75\% AUC), which is the most important task with 4 classes of mental disorders including depression, bipolar, Schizophrenia, and control group. It is also the most complicated task due to our low resource and highly imbalanced data.

\begin{table*}[!t]
\centering
\caption{AUC (\%) on a 25 fold cross-validation for document-level and psychological spectrum predictions using offspring data.}
\label{tbl_auc_offspring}
\setlength\tabcolsep{4pt}
\makebox[\textwidth][c]{
\scalebox{0.7}{
\begin{tabular}{l|ccc|ccccccc}
\toprule
& \multicolumn{3}{|c|}{\textbf{Document-level tasks}} & \multicolumn{7}{|c}{\textbf{Cognitive and Psychological tasks}} \\
\midrule
 & \rot{\textbf{affect}} & \rot{\textbf{coherence}} & \rot{\textbf{richness}} & \rot{\textbf{Spectrum}} & \rot{\textbf{Introvert}} & \rot{\textbf{Extrovert}} & \rot{\textbf{ADHD}} & \rot{\textbf{Anxiety}} & \rot{\textbf{Depression}} & \rot{\textbf{Mood}} \\
\midrule
\textbf{CLAP-LAION \cite{wu2023large_clapstyle}} & 64.23 & 58.48 & 60.81 & 52.67 & 59.14 & 55.26 & 54.35 & 54.17 & 61.74 & 60.56 \\
\midrule
\textbf{TinyBERT+whisperM (Baseline: Transfer)} & 79.99 & 76.09 & 73.23 & 63.84 & 64.00 & 56.09 & 54.15 & 57.60 & 63.31 & 71.90 \\
\textbf{TinyBERT+whisperM (Transfer + MoE)} & 80.82 & 75.98 & 74.03 & 64.29 & 65.18 & 58.59 & 52.63 & 56.39 & 62.73 & 73.36 \\
\textbf{TinyBERT+whisperM (Contrastive + MoE)} & 81.94 & 78.61 & 75.05 & 64.87 & 66.61 & 58.70 & 55.13 & 59.31 & 66.13 & 73.81 \\
\textbf{TinyBERT+whisperM (HILBERT: DualC + MoE)} & \textbf{83.85} & \textbf{79.80} & 76.96 & \textbf{67.33} & \textbf{68.42} & 60.50 & 55.97 & 60.20 & 66.06 & 77.83 \\
\midrule
\textbf{aMiniLM12+spec (Baseline: Transfer)} & 70.22 & 59.47 & 61.66 & 52.71 & 57.85 & 63.47 & 60.95 & 69.93 & 70.70 & 77.10 \\
\textbf{aMiniLM12+spec (Transfer + MoE)} & 71.83 & 60.44 & 62.29 & 54.02 & 62.20 & 64.70 & 62.56 & 70.90 & 72.55 & 79.17 \\
\textbf{aMiniLM12+spec (Contrastive + MoE)} & 73.04 & 62.29 & 63.84 & 56.83 & 61.80 & 66.38 & 64.26 & 72.05 & 73.92 & 80.77 \\
\textbf{aMiniLM12+spec (HILBERT: DualC + MoE)} & 73.17 & 63.14 & 64.66 & 59.74 & 63.17 & 69.75 & \textbf{65.13} & 74.16 & \textbf{74.58} & \textbf{85.31} \\
\midrule
\textbf{sT5XL+whisperM (Baseline: Transfer)} & 79.52 & 73.46 & 75.06 & 50.12 & 57.42 & 67.17 & 56.38 & 56.24 & 63.71 & 64.32 \\
\textbf{sT5XL+whisperM (Transfer + MoE)} & 78.09 & 74.68 & 77.29 & 53.21 & 59.94 & 67.21 & 57.13 & 56.39 & 64.90 & 66.16 \\
\textbf{sT5XL+whisperM (Contrastive + MoE)} & 80.60 & 76.06 & 77.40 & 53.80 & 60.31 & 69.47 & 59.82 & 58.30 & 66.65 & 66.98 \\
\textbf{sT5XL+whisperM (HILBERT: DualC + MoE)} & 80.81 & 78.21 & \textbf{79.97} & 56.09 & 61.82 & 71.05 & 61.63 & 59.86 & 69.65 & 71.60 \\
\midrule
\textbf{nDRoBERTa+w2v2LgFT (Baseline: Transfer)} & 72.42 & 73.18 & 74.11 & 48.98 & 62.58 & 67.85 & 51.99 & 53.95 & 67.57 & 62.06 \\
\textbf{nDRoBERTa+w2v2LgFT (Transfer + MoE)} & 72.04 & 73.49 & 74.58 & 50.15 & 61.53 & 68.06 & 53.14 & 56.38 & 68.21 & 61.31 \\
\textbf{nDRoBERTa+w2v2LgFT (Contrastive + MoE)} & 75.36 & 73.91 & 76.52 & 52.33 & 64.24 & 71.35 & 54.20 & 58.11 & 71.17 & 64.28 \\
\textbf{nDRoBERTa+w2v2LgFT (HILBERT: DualC + MoE)} & 76.39 & 73.88 & 78.16 & 57.43 & 64.95 & \textbf{75.23} & 58.13 & 59.14 & 73.66 & 68.15 \\
\midrule
\textbf{nRoBERTa+w2v2LgFT (Baseline: Transfer)} & 60.50 & 70.63 & 66.84 & 53.24 & 56.51 & 59.35 & 56.13 & 72.36 & 63.32 & 70.63 \\
\textbf{nRoBERTa+w2v2LgFT (Transfer + MoE)} & 60.14 & 72.58 & 67.31 & 52.76 & 58.22 & 60.35 & 54.71 & 71.96 & 63.40 & 73.45 \\
\textbf{nRoBERTa+w2v2LgFT (Contrastive + MoE)} & 62.24 & 72.88 & 69.34 & 55.91 & 60.24 & 62.40 & 58.94 & 73.81 & 65.41 & 74.62 \\
\textbf{nRoBERTa+w2v2LgFT (HILBERT: DualC + MoE)} & 63.50 & 74.37 & 70.33 & 56.34 & 61.15 & 62.56 & 60.73 & \textbf{74.56} & 67.46 & 74.35 \\
\midrule
\textbf{aMiniLM12+whisperM (Baseline: Transfer)} & 77.68 & 73.93 & 73.86 & 50.51 & 61.07 & 55.11 & 53.81 & 61.83 & 60.99 & 61.41 \\
\textbf{aMiniLM12+whisperM (Transfer + MoE)} & 79.65 & 75.89 & 74.83 & 49.56 & 63.06 & 54.92 & 55.18 & 62.69 & 61.47 & 61.00 \\
\textbf{aMiniLM12+whisperM (Contrastive + MoE)} & 79.79 & 76.68 & 76.87 & 52.36 & 62.54 & 56.67 & 56.97 & 64.51 & 63.09 & 62.93 \\
\textbf{aMiniLM12+whisperM (HILBERT: DualC + MoE)} & 81.97 & 79.51 & 76.03 & 54.60 & 65.89 & 56.85 & 60.89 & 66.80 & 63.92 & 64.61 \\
\bottomrule
\end{tabular}}
}
\end{table*}

Table \ref{tbl_auc_offspring} presents AUC (\%) results from 25-fold cross-validation on offspring data across 3 document-level and 7 psychological spectrum tasks. Similar to the parent setting, the full HILBERT (Dual Contrastive + MoE) framework consistently outperforms transfer-only and partial ablation variants across nearly all backbone combinations. Incorporating MoE yields steady improvements over plain transfer learning, while contrastive learning further enhances performance, particularly on psychological traits. The strongest gains are observed in document-level tasks such as affect, coherence, and richness, where HILBERT achieves substantial margins over CLAP-based pre-training, indicating that task-aware multimodal contrastive learning remains effective despite the increased variability in offspring speech.

Across architectures, TinyBERT+WhisperM, aMiniLM12, and larger text encoders (e.g., sT5-XL, nDRoBERTa) benefit substantially from HILBERT, achieving the highest AUCs on several tasks. Notably, aMiniLM12+spec (HILBERT) attains the best performance on multiple psychological outcomes, including ADHD, Depression, and Mood, while TinyBERT+WhisperM (HILBERT) achieves the strongest results on affect and coherence. On the `spectrum' task, HILBERT achieves 67.33\% AUC with TinyBERT and Whisper backbone on offspring data. These trends suggest that audio cues play a more prominent role in offspring data, where prosody, intonation, and pauses provide complementary information beyond lexical content. Performance improvements are broad rather than task-specific, suggesting improved generalization under dual-contrastive supervision. Overall, these results confirm that the advantages of HILBERT extend robustly to offspring data, even under noisier acoustic conditions and greater inter-speaker variability.

\subsubsection{\textbf{Ablation Studies of the Auxiliary Loss Functions:}}
Our final objective incorporates three auxiliary losses: contrastive, CKA, and mutual information (MI). We conduct a full brute force ablation over all non-empty combinations of these losses ($\sum_{r=1}^{N} \binom{N}{r} = 2^N - 1$ where $N=3$) to assess the contribution of each component. The study is performed using the paraTinyBERT and hubertLargeFineTune configuration, which yields the strongest average performance on parent data. Table \ref{tbl_loss_ablation} reports, for each auxiliary loss, the best validation AUC achieved across combinations that include the loss versus those that exclude it. All three losses provide consistent gains. The contrastive loss yields the largest improvement in classification performance. The CKA loss substantially enhances multimodal representation alignment. The MI loss further stabilizes training by balancing modality contributions. Based on these observations, we apply distinct weighting coefficients ($\lambda$) to appropriately emphasize each auxiliary loss in the final objective.
\begin{table}[!t] 
    \centering 
    \caption{Across 7 combinations, each row shows the best AUC (\%) of all combinations that included the auxiliary loss \checkmark vs. those that did not \xmark. Validation AUC (\%) is the average of 5 folds classification on parent data.}
    \label{tbl_loss_ablation}
    \scalebox{0.9}{
    \begin{tabular}{lccccc} 
        \toprule
         & \multicolumn{2}{c}{\textbf{Validation}} \\
        \textbf{Auxiliary Loss} & \textbf{\xmark} & \textbf{\cmark} \\
        \midrule
        \textbf{Contrastive} & 65.8 & 66.1 \\ 
        \textbf{CKA} & 65.7 & 65.9 \\
        \textbf{MI} & 65.7 & 65.8 \\
        \bottomrule
\end{tabular} 
}
\end{table}

\section{Conclusion}
\label{sec_conclusion}
In this work, we introduced HILBERT, a novel framework that effectively addresses the challenges of multimodal audio-text representation learning. Our approach successfully tackles the fundamental problem of aligning cross-modal information while preserving modality-specific characteristics, particularly focusing on the dimensional imbalance between audio and text representations.

The experimental results demonstrate the effectiveness of HILBERT across multiple document-level and psychological spectrum tasks. We achieved an AUC score of 66.75\% and 67.33\% on the `spectrum' task using parent and offspring data, respectively. The `spectrum' task which is the mental disorder prediction is a highly imbalanced multi-class problem and is the most challenging task overall. Interestingly, our findings reveal that the audio component has a more significant impact on offspring data compared to parent data, likely because audio features such as emotions, intonations, and pauses provide additional information beyond spoken words in capturing coherence and mental state in offspring participants.

Several key observations emerge from our experiments. First, the choice of audio and text models significantly influences performance across different tasks, with hubert and whisper showing particular strength in parent and offspring data respectively. Second, smaller language models often outperformed larger models, suggesting that more compact models may be better suited for limited sample sizes in mental health applications.

HILBERT's dual contrastive learning approach, coupled with specialized loss functions and an MoE architecture, offers several advantages over traditional methods. By simultaneously enforcing both inter-modal alignment and intra-modal consistency, our framework creates more balanced and semantically meaningful joint representations. The incorporation of CKA and MI losses ensures that both modalities contribute equitably to the final representation while preserving information-theoretic relevance. Additionally, our multimodal joint encoder effectively captures complex dependencies between modalities through cross-modal self-attention mechanisms, resulting in high-quality document-level representations.

In conclusion, HILBERT advances the state-of-the-art in multimodal representation learning for long sequences, particularly in the context of mental health prediction tasks. Our framework's ability to effectively balance modality contributions while preserving both shared and modality-specific features makes it well-suited for a wide range of downstream applications requiring nuanced understanding of multimodal data.

\bibliographystyle{splncs04}
\bibliography{kdd2026_DCLR}

\end{document}